\documentclass{article}

\usepackage{PRIMEarxiv}

\usepackage[utf8]{inputenc} 
\usepackage[T1]{fontenc}    
\usepackage{hyperref}       
\usepackage{url}            
\usepackage{booktabs}       
\usepackage{nicefrac}       
\usepackage{microtype}      
\usepackage{lipsum}
\usepackage{fancyhdr}       
\usepackage{graphicx}       
\graphicspath{{media/}}     

\usepackage{placeins}
\usepackage{amsmath,amssymb,amsfonts}
\usepackage{breqn}
\usepackage{subcaption} 

\title{A transformer-based deep reinforcement learning approach to spatial navigation in a partially observable Morris Water Maze
}

\author{
  Marte Eggen \\
  Department of Computer Science  \\
  Norwegian University of Science and Technology \\
  Trondheim, Norway\\
  \texttt{marte.eggen@ntnu.no} \\
   \And
  Inga Strümke \\
  Department of Computer Science \\
  Norwegian University of Science and Technology \\
  Trondheim, Norway\\
  \texttt{inga.strumke@ntnu.no} \\
}

\begin{document}
\maketitle

\begin{abstract}
Navigation is a fundamental cognitive skill extensively studied in neuroscientific experiments and has lately gained substantial interest in artificial intelligence research. Recreating the task solved by rodents in the well-established Morris Water Maze (MWM) experiment, this work applies a transformer-based architecture using deep reinforcement learning -- an approach previously unexplored in this context -- to navigate a 2D version of the maze. Specifically, the agent leverages a decoder-only transformer architecture serving as a deep Q-network performing effective decision making in the partially observable environment. We demonstrate that the proposed architecture enables the agent to efficiently learn spatial navigation strategies, overcoming challenges associated with a limited field of vision, corresponding to the visual information available to a rodent in the MWM. Demonstrating the potential of transformer-based models for enhancing navigation performance in partially observable environments, this work suggests promising avenues for future research in artificial agents whose behavior resembles that of biological agents. Finally, the flexibility of the transformer architecture in supporting varying input sequence lengths opens opportunities for gaining increased understanding of the artificial agent's inner representation of the environment.
\end{abstract}


\section{Introduction}
Navigation is a fundamental cognitive skill extensively studied in neuroscientific experiments such as the Morris Water Maze (MWM), a well-established behavioral test for assessing spatial learning in rodents \cite{morris1980, morris1982, morris1984}. In this task, a rodent is placed in a circular pool filled with opaque water and tasked with locating a hidden, submerged platform. Over the course of repeated trials, the rodent learns to navigate to the platform using distal visual cues \cite{morris1984}. In artificial intelligence (AI) research, solving navigation tasks has lately gained substantial interest, specifically in the exploration of replicating neural firing patterns found in parahippocampal areas such as the entorhinal cortex \cite{banino2018}. Particularly, recent work has investigated the navigation strategies of a deep reinforcement learning agent in a \emph{square} version of the MWM task \cite{liu2024}. This study employs an actor-critic network where a shared gated recurrent unit (GRU) is followed by two parallel sets of fully connected layers, providing the policy and value estimate of the current state. The proximal policy optimization algorithm is adapted for training, with auxiliary tasks integrated to improve the agent's performance. The study showed that a recurrent reinforcement learning agent indeed can learn to locate the hidden platform in the simulated MWM, and that the learning process benefited from introducing auxiliary tasks not directly related to the main navigation objective. 

In order to realistically recreate neuroscientific experiments with artificial agents representing living organisms, the environment should be partially observable, to reflect rodents' limited field of vision. However, agents operating with incomplete information often lead to aliasing of multiple states into the same observation \cite{esslinger2022}.
To address this challenge, which generally occurs in partially observable environments, recurrence has commonly been incorporated through GRU and long short-term memory units, to retain historical context from previous steps \cite{liu2024,hausknecht2015}. In recent years, the transformer architecture has demonstrated exceptional success in handling sequential data, outperforming other neural network architectures, with advancements particularly prominent in the field of natural language processing (NLP) \cite{vaswani2017}. Utilizing the transformer model's capacity for sequential processing, \cite{esslinger2022} propose the Deep Transformer Q-Network for partially observable domains, leveraging a decoder-only transformer-based architecture to predict Q-values at each time step. Moreover, previous work has shown that transformers with recurrent position encoding learn spatial representations found in the hippocampus and entorhinal cortex of rodents, also demonstrating the mathematical similarity between transformers and neuroscientific models of the hippocampus \cite{whittington2022}. 

In this work, we apply a transformer-based agent architecture serving as a deep Q-network to represent a rodent operating in a simulated \emph{circular} MWM, corresponding to the experimental design in \cite{morris1984}. To summarize, our work makes the following novel contributions:
\begin{itemize}
    \item An application of a transformer-based architecture in an artificial life related environment. 
    \item A simulation of a circular version of the MWM.
    \item An artificial agent reflecting a rodent's visual experience in a navigation experiment, successfully solving the task without the need of auxiliary tasks for efficient learning.  
\end{itemize}

\section{Background} 

\subsection{The transformer architecture}

The transformer is a deep neural network architecture, first introduced in \cite{vaswani2017} for sequence-to-sequence NLP modeling. The model leverages an attention mechanism to effectively capture long-range dependencies and correlations within sequential data. Specifically, self-attention allows each element of a sequence to attend to all other elements, enabling the model to learn higher-order relations across the entire input. The transformer operates on an embedded input sequence, where each embedding is a learned vector representation of dimension $d$ corresponding to an input observation. Positional encodings are added to preserve information about the observations' relative positions. The resulting embedded sequence $X = \{x_1, ..., x_n\} \in \mathbb{R}^{n \times d}$ is projected through three learnable weight matrices $W^Q \in \mathbb{R}^{d \times d_k}$, $W^K \in \mathbb{R}^{d \times d_k}$, and $W^V \in \mathbb{R}^{d \times d_v}$ to produce queries $Q$, keys $K$ and values $V$, respectively. The attention computation proposed in \cite{vaswani2017}, referred to as scaled dot product attention, is 
\begin{align}
\begin{split}
    Q = XW^Q, \, \,  K &= XW^K, \, \, V = XW^V \\
    \text{Attention}(Q, K, V) 
    &= \text{softmax}
    \left(
    \frac{QK^T}{\sqrt{d_k}}
    \right)
    V \, .
\end{split}
\end{align}
 
The concept of multi-head attention, shown to be more beneficial than a single attention function, computes $h$ attention heads in parallel \cite{vaswani2017}. Each head $i \in \{1, ..., h\}$ uses separate sets of queries, keys, and values, obtained from respective learned linear projections $W_{i}^{Q} \in \mathbb{R}^{d \times d_k}$, $W_{i}^{K} \in \mathbb{R}^{d \times d_k}$, and $W_{i}^{V} \in \mathbb{R}^{d \times d_v}$ with $d_k = d_v = d/h$. The resulting output from all heads is concatenated and followed by a final learned linear transformation $W^{O} \in \mathbb{R}^{hd_v \times d}$. The multi-head attention function is defined as 
\begin{align}
\begin{split}
    \text{MultiHead}(Q, K, V) = \text{Concat}(\text{head}_1, \dots, \text{head}_h)W^{O} \\
    \text{where head}_i = \text{Attention}(QW_{i}^{Q}, KW_{i}^{K}, VW_{i}^{V}) \, .
\end{split}
\end{align}

The transformer model in \cite{vaswani2017} consists of both an encoder and a decoder, and offers a flexible architecture easily modified to accommodate a variety of machine learning tasks. The Generative Pre-trained Transformer (GPT) family of models represents a specific adaptation, only leveraging the decoder part of the original architecture in an auto-regressive manner \cite{radford2018}. The decoder comprises multiple layers, each containing a masked multi-head self-attention mechanism and a feed-forward network, in addition to layer normalization and residual connections. Masked self-attention ensures that each sequence element attends only to previous elements, maintaining the model's auto-regressive property. Among other modifications from the original GPT architecture \cite{radford2018} to its successor, GPT-2, the layer normalization has been moved to the input of the attention and feed-forward modules, with an additional layer normalization at the final output \cite{radford2019}. In this work, the reinforcement learning agent is constructed using the same decoder-only transformer architecture as in GPT-2.

\subsection{Deep Q-networks} \label{background:dqn}
Q-learning is a reinforcement learning algorithm enabling an agent to learn the optimal policy for solving a task in a Markov decision process (MDP) defined by the tuple $(\mathcal{S}, \mathcal{A}, P, R)$, consisting of states $\mathcal{S}$, actions $\mathcal{A}$, transition probabilities $P(s' | s, a)$, and a reward function $R : \mathcal{S} \times \mathcal{A} \rightarrow \mathbb{R}$. The objective of Q-learning is to estimate a function \mbox{$Q : \mathcal{S} \times \mathcal{A} \rightarrow \mathbb{R}$}, denoting the expected discounted reward for executing an action $a \in \mathcal{A}$ in state $s \in \mathcal{S}$ and thereafter following the optimal policy \cite{watkins1992}. 

Given a state $s \in \mathcal{S}$ in which the agent selects an action $a \in \mathcal{A}$, receives a reward $r \in \mathbb{R}$ and leaves the environment in a new state $s' \in \mathcal{S}$, the Q-function is updated according to

\begin{dmath}
    Q(s, a) \leftarrow Q(s, a) + \alpha \left(r + \gamma \max_{a'} Q(s', a') - Q(s, a) \right),
\end{dmath}
where $\alpha$ is the learning rate and $\gamma \in [0, 1)$ is the discount factor. In a deep Q-network, a deep neural network parameterized by $\theta$ approximates the Q-function \cite{mnih2015}. The Q-network is trained by adjusting the parameters $\theta_i$ at iteration $i$ to minimize the mean-squared Bellman error, 

\begin{dmath}
    L_i(\theta_i) = \mathbb{E}_{(s,a,r,s') \sim \mathcal{D}} \left[ \left(r + \gamma \max_{a'} Q(s', a'; \theta_i^{-}) - Q(s, a; \theta_i ) \right)^{2} \right],
    \label{eq:dqn_loss}
\end{dmath}

where the target value uses parameters $\theta_i^{-}$ from a previous iteration, and $\mathcal{D}$ is the pool of stored samples of past transitions $(s, a, r, s')$ \cite{mnih2015}.

\section{Method}

\begin{figure}[t!]
    \centering
    \begin{subfigure}{0.49\textwidth}
        \centering
        \includegraphics[width=\linewidth]{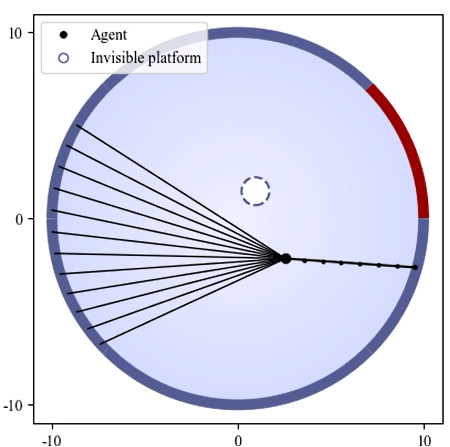}
        \caption{}
    \end{subfigure}
    \begin{subfigure}{0.49\linewidth}
        \centering
        \includegraphics[width=\linewidth]{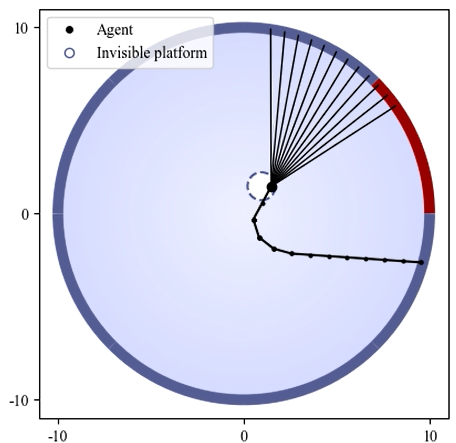}
        \caption{}
    \end{subfigure}
    \caption{The simulated 2D MWM environment includes a visual cue, represented by the red portion of the circumference, and the agent's field of vision comprises 12 distinct sight lines spanning a fixed angle. The navigation objective is to reach the hidden circular platform. The scenarios illustrate (a) the trained agent initially following a direct trajectory towards the center and (b) subsequently rotating until observing the visual cue to adjust the navigation path towards the platform. The smaller dots indicate the steps taken, whereas the larger dot at the front of the line represents the agent's current position.}
    \label{fig:mwm_env}
\end{figure}

In accordance with \cite{liu2024}, the MWM environment is modeled as a partially observable MDP (POMDP), represented by the tuple $(\mathcal{S}, \mathcal{A}, P, R, \Omega, \mathcal{O})$. The distinction between the POMDP and the MDP from section \ref{background:dqn} is that, at time step $t$, the agent receives a partial observation $o_t \in \Omega$ determined by $\mathcal{O} : \mathcal{S} \rightarrow \Omega$ when in a state $s_t \in \mathcal{S}$. The hidden platform requires the agent to navigate using visual cues, which in our simulated version of the MWM are represented by a fixed segment of the circumference in a contrasting color to the rest of the environment. Similar to \cite{liu2024}, our agent is provided with 12 evenly spaced sight lines spanning a fixed 1-radian field of view, aligned with the agent’s facing direction (refer to Fig.~\ref{fig:mwm_env}). Each sight line is represented by two numbers: the distance to the intersecting environment boundary and the associated color, encoded as a unique integer. This results in an observation $\boldsymbol{o}_t \in \mathbb{R}^{24}$ at each time step $t$. 

The agent's experiences are stored in a replay buffer with a maximum size of $50,000$, where the oldest element is removed each time a new element is inserted after the buffer reaches its capacity. The neural network input is a sequence of $n$ consecutive observations $o_{t:t+n} = \{ o_t, ..., o_{t+n} \}$, each embedded through a learned linear transformation into the dimension $d = 128$, and subsequently added to a learned positional encoding. The agent architecture is a stacked decoder-only transformer, equivalent to the structure of \mbox{GPT-2}~\cite{radford2019}, with 2 layers and 8 attention heads, following the hyperparameters in \cite{esslinger2022}. The output from the final decoder layer is projected through a linear layer dimensioned for the action space to yield Q-values, akin to the proposed deep transformer Q-network in \cite{esslinger2022}. The model generates a Q-value per action for each observation in an input sequence, which, at each time step, relies exclusively on preceding observations via the masked self-attention mechanism. This approach diverges from that of \cite{liu2024} by removing the need for recurrence in the agent architecture.  

During evaluation, the selected action corresponds to the highest Q-value from the most recent time step in the input sequence, representing the current state. However, training with intermediate Q-values has been found to enable an efficient training regime and faster learning \cite{esslinger2022}. Following \cite{esslinger2022}, the loss associated with an input sequence is computed by summing the individual losses according to \eqref{eq:dqn_loss} for every sequence element. 

The action space comprises four actions: no action, a forward movement of 1 unit, and a left or right rotation of the agent’s faced angle by 0.2 radians. The circular environment has a radius of 10 units, and a landmark, in the form of a different wall color, covers $1/8$ of the total circumference. The maze contains a hidden platform with a radius of 0.75 units, fixed at a randomly selected position within a 5-unit radius. An episode terminates when the agent successfully reaches the platform, or after 500 steps. At the beginning of each episode, the agent is randomly positioned along the environment boundary, facing the center. This is equivalent to positioning the agent in the same location in each episode while rotating the environment. During training, the agent receives a reward of 1 for locating the platform, $-0.3$ for colliding with the environment boundary, and $-0.0003$ per step, motivating goal-directed behavior. 

The training procedure employs the epsilon-greedy strategy with an exploration rate starting at a value of 0.95 which decays to 0.05 over $10,000$ training steps. During training, the transformer dropout probability is 0.4 to prevent overfitting, the learning rate is 0.0001, the discount factor in \eqref{eq:dqn_loss} is 0.99, and the batch size is 64. A target network is updated every $10,000$ training steps to ensure stable learning, see \eqref{eq:dqn_loss}. 

The design of both the environment and the agent architecture represents a simulation of rodents' navigating behavior observed in neuroscientific experiments. The sequential processing of past observations resembles cognitive processes similar to task-relevant memory mechanisms in the brain, while the agent's limited field of vision reflects that of rodents. Furthermore, the selection of actions available to the artificial agent allows for movements mimicking rodents' physical motion. The epsilon-greedy strategy balances exploration and exploitation by selecting random actions with a certain probability, aligning with the gradual development of rodents' goal-directed strategies. 

The slight negative reward for each step reflects the rodents' natural drive to minimize time spent searching for the platform, while a higher punishment for colliding with the environment boundary is included to avoid unrealistic behavior patterns. Importantly, no auxiliary tasks or supplementary rewards are provided to guide or assist the agent in learning to navigate in the environment or locating the platform, thus preventing any enhancements of performance achieved through artificial means. Despite the absence of auxiliary tasks, the agent demonstrates strong performance \mbox{(refer to section \ref{sec:results_discussion}).} 

\section{Results and analysis} \label{sec:results_discussion}

\begin{figure}[t]
    \centering
    \begin{subfigure}{0.49\linewidth}
        \centering
        \includegraphics[width=\linewidth]{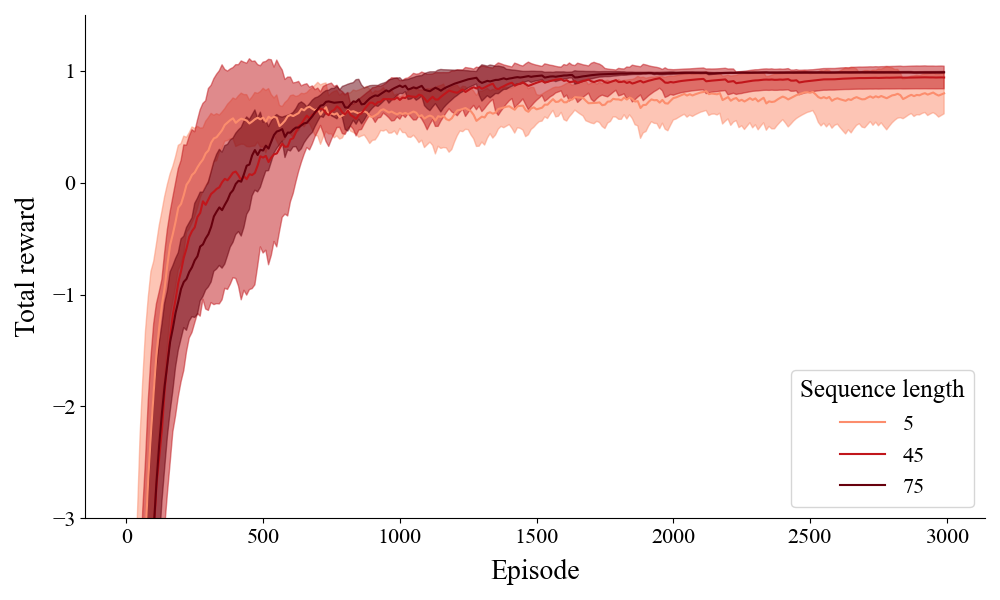}
        \caption{}
        \label{fig:total_reward_v1}
    \end{subfigure}
    \begin{subfigure}{0.49\linewidth}
        \centering
        \includegraphics[width=\linewidth]{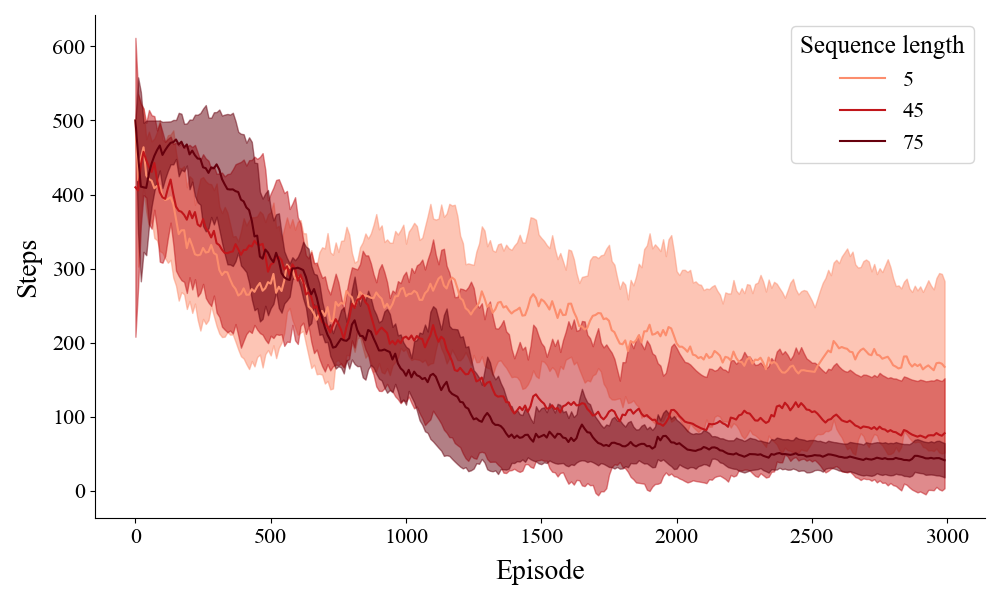}
        \caption{}
        \label{fig:steps_v1}
    \end{subfigure}
    \caption{Exponential moving average of (a) the total reward and (b) the number of steps per episode across five independent runs for input sequence lengths of 5, 45, and 75 observations.}
\end{figure}

The agent was trained for $3,000$ episodes, varying the input sequence lengths to assess performance given different amounts of historical information about previous steps. Fig.~\ref{fig:total_reward_v1} illustrates the total reward per episode, with the associated standard deviation computed across five independent runs, each corresponding to a different platform location. Similarly, Fig.~\ref{fig:steps_v1} presents the number of steps per episode with standard deviations for the same sequence lengths. The figure includes a selection of the tested sequence lengths, showing that longer sequences tend to yield more efficient navigation. However, we observed diminishing returns in performance improvements beyond a sequence length of approximately 45 observations. 

To further investigate navigation strategies, we analyzed the trained agent's behavior using a sequence length of 45 observations over 100 episodes. The agent developed a consistent navigation pattern, where it typically proceeds in a direct trajectory toward the center of the environment, and subsequently rotates until detecting the visual cue to adjust the navigation path towards the platform, as exemplified in Fig.~\ref{fig:mwm_env}. 
While we observe the described behavior in the majority of cases, we find that in a small subset of cases, the agent gets stuck in a repetitive oscillating movement. In these rare instances, observed in approximately 10\% of the cases, the agent proceeds to rotate back and forth, unable to utilize the input sequences of past observations to establish an effective movement pattern. 
\FloatBarrier

\section{Discussion}
The capability of sequential processing is a particularly prominent feature of the transformer architecture, as demonstrated by our agent's efficient learning. The transformer architecture's ability to handle sequential input provides the agent with a history of previous observations, removing the need for recurrence in the network architecture. While we demonstrate the aptness of this architecture and input for solving the navigation task, the unexpected oscillating behavior suggests that the sequential processing capabilities are not fully exploited by our model configuration, or even the GPT-2 architecture. Thus, an interesting direction for further study is to understand how different aspects of the task are internalized with varying sequence lengths. 

Furthermore, we aim to utilize techniques from the field of explainable AI (XAI), suited for transformer architectures, to gain a better understanding of the agent's inner representations, i.a.\ how the agent internalizes information about the environment and solution strategies. While our study is limited to visual inspection of the agent's navigation strategies, the application of XAI might provide deeper and more detailed insights, primarily regarding the extent to which the agent utilizes the landmark for navigation. Additionally, such analysis could facilitate more precise optimization of the agent's behavior, and the agent's robustness could be assessed through more challenging platform locations. Finally, it is possible that XAI techniques could inform how different sequence lengths influence the agent's decision making, both in the cases of efficient navigation, and particularly in cases of ``stuck behavior''. 

Another direction for future research involves refining the selection of experiences to add to the replay buffer, which functions as the agent’s selective, task-specific memory. We aim to investigate whether a biology-inspired approach for filling the replay buffer could improve the relevance of the input sequences. Understanding which past experiences are most critical for a rodent's spatial learning could inform the design of a corresponding algorithmic experience selection criterion for artificial agents, potentially improving the learning process.

\section{Conclusion}
This work presents the successful application of a transformer-based architecture serving as a deep Q-network in an environment replicating the circular MWM, a well-established neuroscientific experiment for spatial learning in rodents. 
The simulation is intended to preserve the natural challenges faced by rodents in navigation tasks, ensuring the artificial agent's learning experience aligns with the physical constraints experienced by rodents. The agent operates in a partially observable environment to reflect a rodent's visual experience in navigating the maze, and we find that the transformer architecture is well-suited for this class of decision making tasks.

\bibliographystyle{unsrt}  
\bibliography{references}

\end{document}